\documentclass[10pt,twocolumn,letterpaper]{article}

\usepackage[review]{cvpr}      

\usepackage{multirow} 
\usepackage{colortbl}

\usepackage[dvipsnames]{xcolor}

\definecolor{cvprblue}{rgb}{0.21,0.49,0.74}
\usepackage[pagebackref,breaklinks,colorlinks,citecolor=cvprblue]{hyperref}

\def\paperID{xxxx} 
\def\confName{CVPR}

\title{Skeleton2vec: A Self-supervised Learning Framework with Contextualized Target Representations for Skeleton Sequence}

\author{First Author\\
Institution1\\
Institution1 address\\
{\tt\small firstauthor@i1.org}
\and
Second Author\\
Institution2\\
First line of institution2 address\\
{\tt\small secondauthor@i2.org}
}

\begin{document}
\maketitle
\begin{abstract}
Self-supervised pre-training paradigms have been extensively explored in the field of
skeleton-based action recognition. In particular, methods based on
\textbf{masked prediction} have pushed the performance of pre-training to a new height.
However, these methods take low-level features, such as raw joint coordinates or
temporal motion, as prediction targets for the masked regions, which is suboptimal.
In this paper, we show that using high-level contextualized features as prediction
targets can achieve superior performance. Specifically, we propose \textbf{Skeleton2vec},
a simple and efficient self-supervised 3D action representation learning framework,
which utilizes a transformer-based teacher encoder taking unmasked training samples as
input to create \textbf{latent contextualized representations} as prediction targets.
Benefiting from the self-attention mechanism, the latent representations generated by
the teacher encoder can incorporate the global context of the entire training samples,
leading to a richer training task.
Additionally, considering the high temporal correlations in skeleton sequences, we propose a
\textbf{motion-aware tube masking strategy} which divides the skeleton sequence into
several tubes and performs persistent masking within each tube based on motion priors,
thus forcing the model to build long-range spatio-temporal connections and focus on
action-semantic richer regions. Extensive experiments on NTU-60, NTU-120, and PKU-MMD
datasets demonstrate that our proposed Skeleton2vec outperforms previous methods and
achieves state-of-the-art results.
The source code of Skeleton2vec is available at \url{https://github.com/Ruizhuo-Xu/Skeleton2vec}.
\end{abstract}    
\section{Introduction}
\label{sec:intro}

\begin{figure}[htbp]
	\centering
	\begin{subfigure}{0.49\linewidth}
		\centering
		\includegraphics[width=1.0\linewidth]{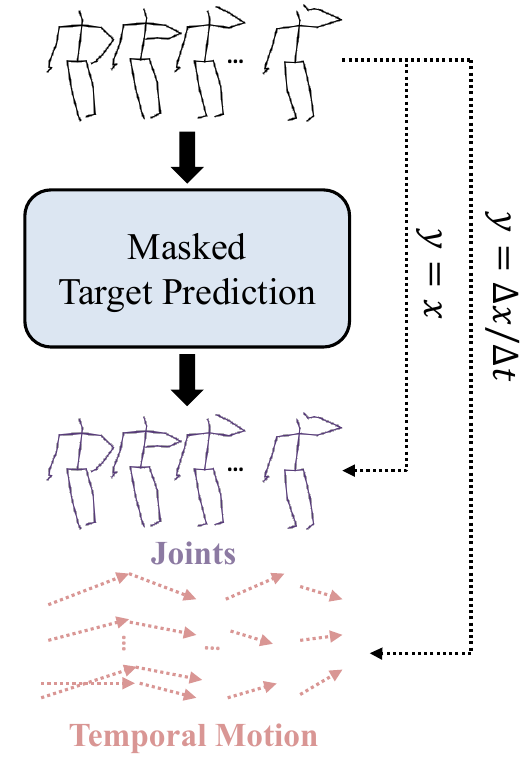}
		\caption{MAE-like}
		\label{fig1:MAE}
	\end{subfigure}
	\centering
	\begin{subfigure}{0.49\linewidth}
		\centering
		\includegraphics[width=1.0\linewidth]{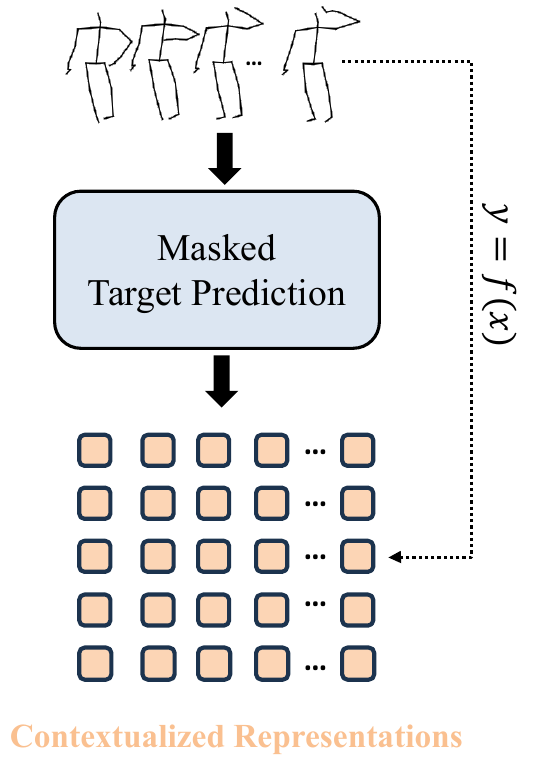}
		\caption{Skeleton2vec(Ours)}
		\label{fig1:skeleton2vec}
	\end{subfigure}
    \caption{
    A comparative illustration of the prediction targets between MAE-like methods (a) and
    ours Skeleton2vec (b). Skeleton2vec utilizes an teacher encoder $f(x)$ to generate
    globally contextualized representations as the prediction targets, instead of
    isolated joints or temporal motion with only local context.
    }
    \label{fig1}
\end{figure}

Human action recognition has significant applications in the real world, such as
security, human-robot interaction, and virtual reality. The development of depth
sensors and advancements in pose estimation algorithms \cite{2018OpenPose, fang2017rmpe, xu2020deep}
have propelled skeleton-based action recognition into a popular research topic,
owing to its computational efficiency, background robustness, and privacy preservation.
A series of fully-supervised skeleton-based human action recognition methods have
been developed using CNNs \cite{du2015skeleton,li2017skeleton}, RNNs \cite{liu2016spatio,zhang2017view},
and GCNs \cite{yan2018spatial,chen2021channel}. Despite their promising performance,
these methods rely on large amounts of manually annotated data, which is expensive,
labor-intensive, and time-consuming to obtain. This circumstance motivates us
to explore self-supervised representation learning for 3D actions.

Earlier works \cite{lin2020ms2l, nie2020unsupervised, su2020predict, zheng2018unsupervised}
have employed various pretext tasks, such as motion prediction, jigsaw puzzle recognition,
and masked reconstruction, to learn 3D action representations. Recently, contrastive
learning methods \cite{rao2021augmented, guo2022contrastive, moliner2022bootstrapped, lin2023actionlet}
have gained prominence. However, these methods often require carefully designed
data augmentations and tend to encourage the encoder to learn more global
representations, thereby neglecting local spatiotemporal information.
With the rise of transformer models \cite{vaswani2017attention}, self-supervised
pre-training methods based on masked prediction tasks have become mainstream in
visual representation learning \cite{rao2021augmented, guo2022contrastive, moliner2022bootstrapped, lin2023actionlet}.
Works like SkeletonMAE \cite{yan2023skeletonmae, wu2023skeletonmae} and MAMP \cite{mao2023masked} have
attempted to transfer MAE \cite{he2022masked} methods to the field of 3D action representation
learning, achieving promising results. However, these MAE-like methods inefficiently
utilize model capacity by focusing on low-level high-frequency details with
raw joint coordinates or temporal motion as learning targets, which is suboptimal
for modeling high-level spatiotemporal structures. We believe that using
higher-level prediction targets will guide the model to learn better representations
and improve pre-training performance.

Motivated by this idea, we propose Skeleton2vec, a simple and efficient self-supervised
framework for 3D action representation learning. Addressing the limitations of existing
MAE-like methods, as illustrated in \cref{fig1}, Skeleton2vec leverages contextualized
prediction targets. Following the work of data2vec \cite{baevski2022data2vec, baevski2023efficient},
we employ a teacher encoder that takes unmasked training samples to generate latent
contextualized representations as targets. We then use a student encoder, taking a
masked version of the sample as input, combined with an asymmetric decoder to
predict data representations at the masked positions. The entire model is based on the
vanilla transformer architecture. The self-attention mechanism ensures that the
constructed targets are contextualized, incorporating information from the entire
sample, making them richer than isolated targets (\eg raw joint coordinates)
or targets based on local context (\eg temporal motion).

Additionally, considering the strong spatiotemporal correlations in 3D skeleton sequences,
we propose a motion-aware tube masking strategy. Initially, we divide the input skeleton
sequence along the temporal axis into multiple tubes, where frames within each tube share
a masking map to avoid information leakage from neighboring frames. This forces the model to
extract information from distant time steps for better prediction. We then guide the sampling
of masked joints based on the spatial motion intensity of body joints within each tube.
Joints with higher motion intensity will be masked with higher probability, allowing
the model to focus more on spatiotemporal regions with rich action semantics. Compared
to random masking, our method better utilizes the spatiotemporal characteristics and
motion priors of 3D skeleton sequences, effectively improving pre-training performance.

In summary, the main contributions of this work are three-fold:
\begin{itemize}
    \item{
        We propose the Skeleton2vec framework, which uses contextualized representations
        from a teacher encoder as prediction targets, enabling the learned representations
        to have stronger semantic associations.
    }
    \item{
        We introduce a motion-aware tube masking strategy that performs persistent masking
        of joints within tubes based on spatial motion intensity, forcing the model to
        build better long-range spatiotemporal connections and focus on more semantic-rich regions.
    }
    \item{
        We validate the effectiveness of our method on three large-scale 3D skeleton-based
        action recognition datasets and achieve state-of-the-art results.
    }
\end{itemize}
\section{Related Work}
\label{sec:related}

\subsection{Self-supervised Skeleton-based Action Recognition}
Previous studies \cite{zheng2018unsupervised,su2020predict,lin2020ms2l} on
self-supervised representation learning for skeleton-based action recognition
utilize various pretext tasks to capture motion context.
For instance, LongTGAN \cite{zheng2018unsupervised} leverages sequence reconstruction
to learn 3D action representations. P\&C \cite{su2020predict} employs a weak
decoder to enhance representation learning. MS2L \cite{lin2020ms2l} employs
motion prediction and jigsaw puzzle tasks. Yang et al. \cite{yang2021skeleton}
introduce a skeleton cloud colorization task.
Contrastive learning methods have gained prominence in 3D action representation learning
\cite{he2020momentum,grill2020bootstrap,rao2021augmented,guo2022contrastive,moliner2022bootstrapped,lin2023actionlet}.
AS-CAL \cite{rao2021augmented} and SkeletonCLR \cite{li20213d} utilize momentum encoder
and propose various data augmentation strategies. AimCLR \cite{guo2022contrastive} introduces
extreme augmentations. ActCLR \cite{lin2023actionlet} performs adaptive action modeling
on different body parts. Despite their remarkable results, contrastive learning methods
often overlook local spatio-temporal information, a crucial aspect for 3D action modeling.

The surge in popularity of transformers has led to the mainstream adoption of self-supervised
pretraining based on masked visual modeling for visual representation learning \cite{he2022masked,bao2021beit}.
SkeletonMAE \cite{wu2023skeletonmae} and MAMP \cite{mao2023masked} apply the Masked Autoencoder (MAE)
approach to 3D action representation learning. SkeletonMAE employs a skeleton-based encoder-decoder
transformer for spatial coordinate reconstruction, while MAMP introduces Masked Motion Prediction
to explicitly model temporal motion. In this study, we demonstrate that utilizing higher-level
contextualized representations as prediction targets for masked regions yields superior performance
compared to directly predicting raw joint coordinates or temporal motion.

\subsection{Masked Image Modeling}
BEiT \cite{bao2021beit} pioneered masked image modeling (MIM) for self-supervised pretraining
of visual models, aiming to recover discrete visual tokens from masked patches.
Subsequently, various prediction targets for MIM have been explored. MAE \cite{he2022masked}
and SimMIM \cite{xie2022simmim} treat MIM as a denoising self-reconstruction task, utilizing
raw pixels as the prediction target. MaskFeat \cite{wei2022masked} replaces pixels with HOG
descriptors to enable more efficient training and achieve superior results. PeCo \cite{dong2023peco}
introduces a perceptual loss during dVAE training to generate semantically richer discrete
visual tokens, surpassing BEiT.
These works demonstrate superior performance by utilizing
higher-level and semantically richer prediction targets in MIM.
To further enhance performance,
data2vec \cite{baevski2022data2vec,baevski2023efficient} employs self-distillation to
leverage latent target representations from the teacher model output at masked positions.
Compared to isolated targets like visual tokens or pixels, these contextualized representations
encompass relevant features from the entire image, enabling improved performance.

In this research, we introduce the data2vec framework into self-supervised pretraining of
skeleton sequences, utilizing latent contextualized target representations from the teacher
model to guide the student model in learning more effective 3D action representations.

\section{Method}
\subsection{Overview}

\begin{figure*}[htbp]
	\centering
	\begin{subfigure}{0.49\linewidth}
		\centering
		\includegraphics[width=0.95\linewidth]{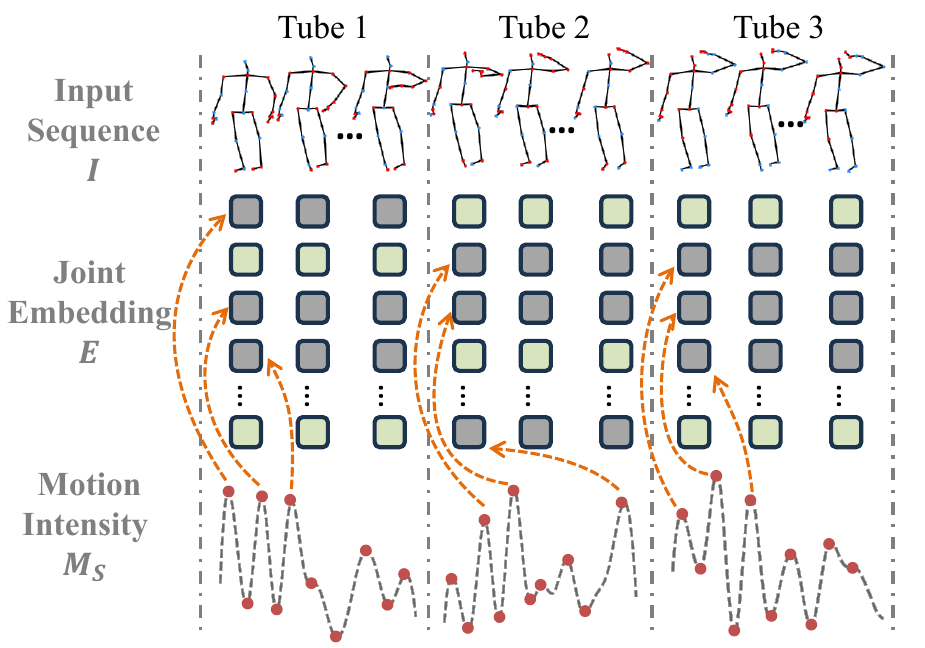}
		\caption{Motion-Aware Tube Masking}
		\label{fig2:motion_aware_tube_masking}
	\end{subfigure}
	\centering
	\begin{subfigure}{0.49\linewidth}
		\centering
		\includegraphics[width=0.95\linewidth]{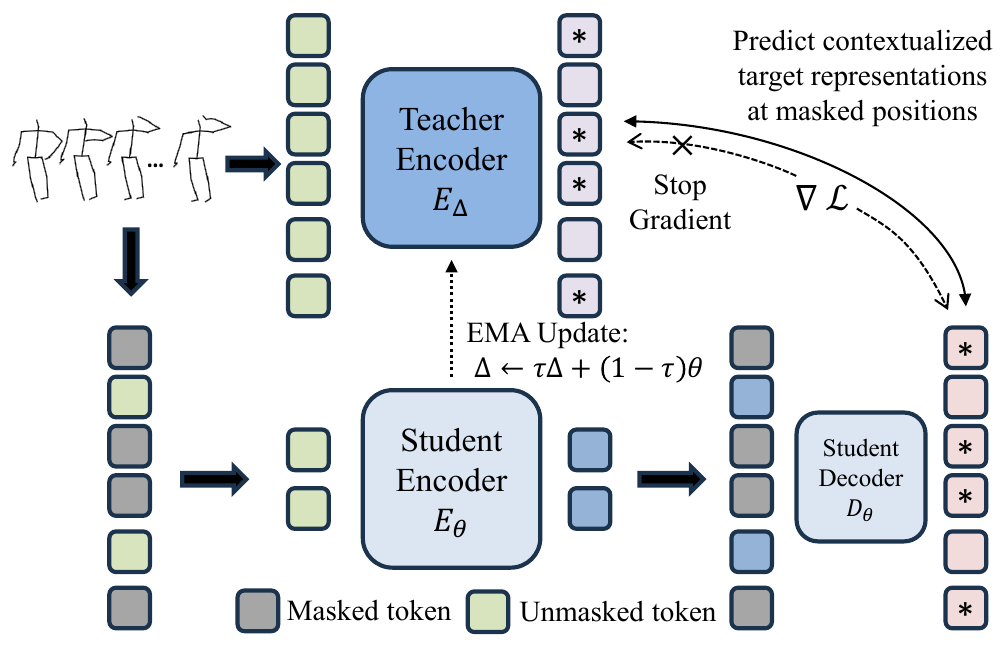}
		\caption{Skeleton2vec}
		\label{fig2:skeleton2vec}
	\end{subfigure}
    \caption{
    The overall pipeline of the proposed Skeleton2vec framework.
    We adopt the motion-aware tube masking strategy (a) to guide the masking process,
    which prevents information leakage between adjacent frames and allows the model
    to focus more on semantically rich regions of motion. Subsequently, the teacher
    encoder $E_{\Delta}$ receives unmasked samples to construct latent contextualized targets,
    while the student encoder $E_{\theta}$ receives masked versions of the samples
    and predicts corresponding representations at the masked positions.
    }
    \label{fig2}
\end{figure*}

The overall framework of Skeleton2vec is shown in \cref{fig2}.
It takes a skeleton sequence $I \in \mathbb{R}^{T_{s} \times V \times C_{s}}$ as
input, where $T_{s}$ is the the number of frames,
$V$ is the number of joints, and $C_{s}$ is the
the coordinates of joints.
Similar to most visual transformers \cite{dosovitskiy2020image},
the skeleton sequence is first divided into fixed-size patches and then linearly
transformed into patch embedding $E \in \mathbb{R}^{T_{e} \times V \times C_{e}}$.
After that, we employ the motion-aware tube masking strategy to guide the masking of
joints. The teacher model constructs the full contextualized prediction targets
using unmasked training samples, while the student model receives the masked version
of the samples and predicts corresponding representations at the masked positions.

As our student model, we adopt an asymmetric encoder-decoder architecture, where the
encoder operates solely on non-masked tokens. The lightweight decoder inserts masked
tokens into the latent representations outputted by the encoder, forming a full set
for predicting the targets.
The teacher encoder shares the same model structure as the student. After accomplishing
the aforementioned pre-training task, the teacher encoder is retained for downstream task fine-tuning.

\subsection{Model Architecture}
\noindent \textbf{Encoder:}
Following MAMP \cite{mao2023masked}, we first divide the raw skeleton sequence
$I \in \mathbb{R}^{T_{s} \times V \times C_{s}}$ into non-overlapping segments
$I' \in \mathbb{R}^{T_{e} \times V \times (l \cdot C_{s})}$, where $T_{e}=T_{s}/l$
and $l$ is the length of each segment.
A trainable linear projection is then applied to each joint to obtain the embedding:
\begin{equation}
    \label{eq:joint_embedding_1}
    E_{j} = \text{LinearProj}(I') \in \mathbb{R}^{T_{e} \times V \times C_{e}},
\end{equation}
where $C_{e}$ represents the dimension of the embedding.
Temporal positional embedding $E_{t} \in \mathbb{R}^{T_{e} \times 1 \times C_{e}}$
and spatial positional embedding $E_{s} \in \mathbb{R}^{1 \times V \times C_{e}}$
are then added to the joint embedding to yield the final input:
\begin{equation}
    \label{eq:joint_embedding_2}
    E = E_{j} + E_{t} + E_{s},
\end{equation}

For the teacher encoder, the entire set is flattened as input $E^{T} \in \mathbb{R}^{N_{T} \times C_{e}}$,
where $N_{T}=T_{e} \times V$ represents the total number of tokens in the
skeleton sequence. For the student encoder, most tokens are masked, and
only the unmasked tokens are utilized as input, flattened as $E^{S} \in \mathbb{R}^{N_{S} \times C_{e}}$,
where $N_{S}=T_{e} \times V \times (1-m)$ denotes the number of visible tokens,
and $m$ is the masking ratio. 
Subsequently, $L_{e}$ layers of vanilla transformer blocks are applied to extract
latent representations. Each block comprises a multi-head self-attention (MSA)
module and a feed-forward network (FFN) module. Residual connections are employed
within each module, followed by layer normalization (LN).

\noindent \textbf{Decoder:} The decoder input
$D \in \mathbb{R}^{T_{e} \times V \times C_{e}}$ contains the full set of tokens,
including the latent representations of visible encoded tokens $Z^{S}_{e}$ and the inserted masked tokens.
Each masked token is represented by a shared learnable vector $E_{M} \in \mathbb{R}^{C_{e}}$,
indicating missing information to be predicted at that position. Similar to the
encoder, spatial positional embedding $E'_{s}$ and temporal positional embedding
$E'_{t}$ are added to all tokens to assist masked tokens in locating their positions.
The decoder employs an additional $L_{d}$ layers of transformer blocks for masked prediction.

\subsection{Contextualized Target Prediction}
Rather than relying on isolated raw joints or temporal motion with limited local context, we employ a
transformer-based teacher encoder to construct globally contextualized prediction targets,
thereby introducing a diverse training task.

\noindent \textbf{Contextualized Target Representations:}
We extract features from the output of each FFN block in every layer of the
teacher encoder and average them to form our training targets. Following data2vec 2.0
\cite{baevski2023efficient}, the features from each layer are normalized with
instance normalization \cite{ulyanov2016instance} before averaging.
Finally, the averaged features are normalized by layer normalization to serve as
the prediction targets. Normalizing the targets helps prevent the model from
collapsing to a trivial solution, and also prevents any single layer's features
from dominating. The generation of the target representations can be formulated as:
\begin{equation}
    \label{eq:target_rep}
    \begin{aligned}
        Y' &= \frac{1}{L_e}\sum_{l=1}^{L_e} \text{IN}(Z_l^T), \\
        Y &= \text{LN}(Y'),
    \end{aligned}
\end{equation}
where IN and LN refer to instance normalization and layer normalization, respectively.
$Z_l^T$ denotes the output of the FFN block in the $l^{th}$ layer of the teacher encoder.

\noindent \textbf{Target Prediction:}
Given the output $H_d$ of the student decoder, we employ an additional linear prediction head to
regress the contextualized target representations of the teacher:  
\begin{equation}
    \label{eq:target_pred}
    \hat{Y} = \text{LinearPred}(H_d),
\end{equation}

Finally, we adopt L2 loss as our learning objective, calculating loss only for the
masked positions:
\begin{equation}
    \label{eq:loss}
    \mathcal{L} = \frac{1}{|\mathcal{M}|}\sum_{i \in \mathcal{M}}||Y_i - \hat{Y}_i||_2^2,
\end{equation}
where $\mathcal{M}$ denotes the set of masked positions.

\noindent \textbf{Teacher Parameterization:}
The student model weights $\theta$ are updated through backpropagation on the loss
gradients. The teacher model weights $\Delta$ are initialized to be the same as the
student weights and parameterized during training by taking an exponentially moving
average (EMA) of the student weights:  
\begin{equation}
    \label{eq:ema}
    \Delta \leftarrow \tau\Delta + (1-\tau)\theta,
\end{equation}
where $\tau$ is a hyperparameter controlling the update frequency of the teacher
weights using a linearly increasing schedule, gradually increasing from an initial
value $\tau_0$ to 1 throughout training.

\subsection{Motion-Aware Tube Masking}
\label{sec:motion-aware_tube_masking}
We propose the motion-aware tube masking strategy to address the issue of high
spatiotemporal correlations in skeleton sequences.

\noindent \textbf{Tube Division:}
The tube masking strategy, initially introduced by VideoMAE \cite{tong2022videomae},
considers the entire video sequence along the temporal axis as a single tube,
sharing the same masking map across different frames. This mitigates the information
leakage issue between adjacent frames.
Although the skeleton sequence is derived from the
video, directly applying this single-tube
masking strategy to skeleton data is suboptimal due to the inherent structural differences.
In video data, the basic units for masking are image patches in each frame. Due to scene
motion or camera viewpoint changes, a masked body part like the hand in the first frame
may find its correspondence in unmasked regions in later frames far apart, which facilitates
long-range dependency modeling. In contrast, the basic units for masking in skeleton sequences
are the joints in each skeleton frame, where the same-order joints have explicit correspondence
across frames. As a result, a body part masked in the first skeleton frame will remain
masked in all frames, causing a complete loss of information for that part, which makes
the masked prediction task overly difficult and harms the model's learning capability.
To address this, as illustrated in \cref{fig2:motion_aware_tube_masking},
we empirically divide the skeleton sequence along the time axis into multiple tubes
instead of one tube. Each tube shares the same masking map to force the model to extract
information from farther time steps, while different tubes use different masking maps
to avoid joints being masked throughout. The tube division can be represented as:
\begin{equation}
    \label{eq:deviding_tubes}
    E'=\text{Reshape}(E) \in \mathbb{R}^{N \times \alpha \times V \times C_{e}}, 
\end{equation}

\noindent where $\alpha$ is tube length and $N=\frac{T_{e}}{\alpha}$ is number of tubes.

\noindent \textbf{Motion-Aware Sampling:}
Regions with larger motion intensity intuitively contain richer semantic information
about actions. Therefore, we utilize the spatial motion intensity of each human body
joint within a tube as empirical guidance to generate the masking map.

Specifically, we first extract the corresponding motion sequence
$M \in \mathbb{R}^{T_{s} \times V \times C_{s}}$ from the input skeleton sequence
$I \in \mathbb{R}^{T_{s} \times V \times C_{s}}$ by calculating temporal differences
of corresponding joint coordinates between adjacent frames:
\begin{equation}
    \label{eq:motion}
    M_{i,:,:} = \begin{cases} 
    I_{i+1,:,:}-I_{i,:,:}, & i \in 0, \dots, T_{s}-1\\
    0, & i=T_{s}  
\end{cases}
\end{equation}

Similar to joint embedding in the encoder, we reshape $M$ into non-overlapping
segments $M' \in \mathbb{R}^{T_{e} \times V \times (l \cdot C_{s})}$ to match the
shape of input sequence $I'$. We then calculate the motion intensity of each joint
within a segment as:
\begin{equation}  
    \label{eq:motion_intensity}
    S_{i,:} = \sum_{k=0}^{l\cdot C_{s}}|M'_{i,:,k}| \in \mathbb{R}^{T_{e} \times V}, \quad i=0,\dots,T_{e}
\end{equation}

Afterwards, we compute the spatial motion intensity of each body joint within a tube,
normalizing it along the spatial dimension:  
\begin{equation}
    \label{eq:tube_motion_intensity_norm}
    \begin{aligned}
        T_{i,:} &= \sum_{j=i}^{i+\alpha}S_{j,:} \in \mathbb{R}^{N \times V}, \quad &i=0,\dots,N \\
        T'_{i,:} &= T_{i,:} / \text{max}(T_{i,:}), \quad &i=0,\dots,N  
    \end{aligned}
\end{equation}

Finally, we utilize the normalized spatial motion intensity to generate a unique
masking map for each tube:
\begin{equation} 
    \label{eq:mask_sampling}
    \begin{aligned}
        p &= \eta + \beta \cdot T', \quad &\eta \sim U(0,1) \\  
        \mathcal{M}_{i} &= \text{argsort}(p_{i,:})[-K:], \quad &i=0,\dots,N
    \end{aligned}
\end{equation}

\noindent where $\eta$ is random noise drawn from a uniform distribution between
0 and 1, $\beta$ is a hyperparameter controlling the influence of spatial motion
intensity on sampling, $\mathcal{M}_{i}$ is the masking map for $i^{th}$ tube,
$K=V\times (1-m)$ is the number of joints to be masked, and $m$ is the masking ratio.
By customizing motion-aware masking maps for each tube, the model is encouraged
to focus more on semantically richer regions, leading to improved spatiotemporal representations.

\begin{table*}[htbp]
    \centering
    \begin{tabular}{l c c c c c c}
      \toprule
      \multirow{2}{*}{Method} &
      \multirow{2}{*}{Input} &
      \multicolumn{2}{c}{NTU 60} &
      \multicolumn{2}{c}{NTU 120} &
      PKU II\\
      & & XSub(\%) & XView(\%) & XSub(\%) & XSet(\%) & XSub(\%) \\
      \midrule
      \rowcolor{Gray!20} \multicolumn{7}{l}{\textit{Other pretext tasks:}} \\
      LongTGAN \cite{zheng2018unsupervised} & Single-stream & 39.1 & 48.1 & - & - & 26.0 \\
      P\&C \cite{su2020predict} & Single-stream & 50.7 & 75.3 & 42.7 & 41.7 & 25.5 \\
      \midrule
      \rowcolor{Gray!20} \multicolumn{7}{l}{\textit{Contrastive Learning:}} \\
      CrosSCLR \cite{li20213d} & Three-stream & 77.8 & 83.4 & 67.9 & 66.7 & 21.2 \\
      AimCLR \cite{guo2022contrastive} & Three-stream & 78.9 & 83.8 & 68.2  & 68.8 & 39.5 \\
      CPM \cite{zhang2022contrastive} & Single-stream & 78.7 & 84.9 & 68.7 & 69.6 & - \\
      PSTL \cite{Zhou2023SelfsupervisedAR} & Three-stream & 79.1 & 83.8 & 69.2 & 70.3 & 52.3 \\
      CMD \cite{mao2022cmd} & Single-stream & 79.4 & 86.9 & 70.3 & 71.5 & - \\
      HaLP \cite{shah2023halp} & Single-stream & 79.7 & 86.8 & 71.1 & 72.2 & 43.5 \\
      HiCo-Transformer \cite{hico2023} & Single-stream & 81.1 & 88.6 & 72.8 & 74.1 & 49.4 \\
      SkeAttnCLR \cite{Hua2023SkeAttnCLR} & Three-stream & 82.0 & 86.5 & 77.1 & 80.0 & 55.5 \\
      ActCLR \cite{lin2023actionlet} & Three-stream & 84.3 & 88.8 & 74.3 & 75.7 & - \\
      \midrule
      \rowcolor{Gray!20} \multicolumn{7}{l}{\textit{Masked Prediction:}} \\
      SkeletonMAE \cite{yan2023skeletonmae} & Single-stream & 74.8 & 77.7 & 72.5 & 73.5 & 36.1 \\
      MAMP \cite{mao2023masked} & Single-stream & 84.9 & 89.1 & 78.6 & 79.1 & 53.8 \\
      \textbf{Skeleton2vec(Ours)} & Single-stream & \textbf{85.7} & \textbf{90.3} & \textbf{79.7} & \textbf{81.3} & \textbf{55.6} \\
      \bottomrule
    \end{tabular}
    \caption{
      Performance comparison in linear evaluation protocol on NTU 60, NTU 120,
      and PKU MMD datasets.  \textit{Single-stream} refers to Joint,
      while \textit{Three-stream} denotes Joint+Motion+Bone.
    }
    \label{tab:linear}
\end{table*}

\section{Experiments}
\subsection{Datasets}
We evaluate our method on three large-scale 3D skeleton-based action recognition
datasets: NTU RGB+D 60, NTU RGB+D 120, and PKU Multi-Modality Dataset (PKUMMD).

NTU RGB+D 60 \cite{shahroudy2016ntu} contains 56,880 skeleton sequences across 60
action categories performed by 40 subjects. We follow the recommended cross-subject
and cross-view evaluation protocols. For cross-subject, sequences from 20 subjects
are used for training and the rest are used for testing. For cross-view, training
samples are from cameras 2 and 3, while testing samples are from camera 1. 

NTU RGB+D 120 \cite{liu2019ntu} is an extension of NTU RGB+D 60 with 114,480
skeleton sequences across 120 action categories performed by 106 subjects.
The authors also propose a more challenging cross-setup evaluation protocol, where
sequences are divided into 32 setups based on camera distance and background. Samples
from 16 setups are used for training and the rest are used for testing.

PKUMMD \cite{liu2017pku} contains nearly 20,000 skeleton sequences across 52 action
categories. We adopt the cross-subject protocol, where training and testing sets are
split based on subject ID. PKUMMD consists of two parts: PKU-I and PKU-II. PKU-II is
more challenging due to larger view variations that introduce more skeleton noise.
For PKU-II, there are 5,332 sequences for training and 1,613 for testing.

\subsection{Settings}
\noindent \textbf{Data Processing:}
We employed the data preprocessing method from DG-STGCN \cite{duan2022dg} to apply
uniform sampling to a given skeleton sequence, generating subsequences as training
samples. The number of frames $T_{s}$ for sampling is set to 90.
During the training, we applied random rotation as data augmentation on the
sampled subsequences to enhance robustness against view variation. During the testing,
we averaged the scores of 10 subsequences to predict the class.

\noindent \textbf{Network Architecture:}
We adopted the same network architecture setting as MAMP \cite{mao2023masked}, with the
encoder layers $L_{e}$ set to 8, decoder layers $L_{d}$ set to 3, embedding dimension
set to 256, the number of heads in the multi-head self-attention module set to 8, and
the hidden dimension of the feed-forward network set to 1024. For Joint Embedding,
the length $l$ of each segment is set to 3.

\noindent \textbf{Pre-training:}
In the pre-training, the initial value of the EMA parameter $\tau$ is set to 0.9999.
The masking ratio $m$ of the input sequence is set to 90\%.
The tube length $\alpha$ for motion-aware tube masking is set to 5,
and the sampling parameter $\beta$ is set to 0.1. We utilized the AdamW optimizer
with weight decay of 0.05 and betas (0.9, 0.95). The model was trained for a total of
600 epochs, with the learning rate linearly increasing to 1e-3 during the first 20
warmup epochs, and then decaying to 1e-5 according to a cosine decay schedule.
Our model was trained on 2 RTX 4090 GPUs, with a total batch size of 128.

\subsection{Evaluation and Comparison}
\noindent \textbf{Linear Evaluation:}
In the linear evaluation protocol, the parameters of the pre-trained encoder are
fixed to extract features. A trainable linear classifier is then applied for
classification. We train for 100 epochs in total using SGD optimizer with momentum of
0.9 and batch size of 256.
The initial learning rate is set to 0.1 and is decreased to 0 following a
cosine decay schedule.
Our results are evaluated on three datasets: NTU-60, NTU-120, and PKU-MMD.
Comparison with the latest methods reveals the superiority of our proposed Skeleton2vec,
as illustrated in \cref{tab:linear}. Notably, in contrast to contrastive learning
methods, Skeleton2vec, employing the masked prediction approach, demonstrates significant advantages.
Furthermore, Skeleton2vec outperforms other masked prediction methods across all datasets.
Particularly, on the NTU-60 XView and NTU-120 XSet datasets, Skeleton2vec exhibits
superior performance over the previously state-of-the-art method MAMP by 1.2\% and 2.2\%,
respectively, highlighting the strength of our contextualized prediction targets.

\begin{table*}[htbp]
    \centering
    \begin{tabular}{l c c c c c c}
      \toprule
      \multirow{2}{*}{Method} &
      \multirow{2}{*}{Input} &
      \multirow{2}{*}{Backbone} &
      \multicolumn{2}{c}{NTU 60} &
      \multicolumn{2}{c}{NTU 120} \\
      & & & XSub(\%) & XView(\%) & XSub(\%) & XSet(\%) \\
      \midrule
      \rowcolor{Gray!20} \multicolumn{7}{l}{\textit{Other pretext tasks:}} \\
      Colorization \cite{yang2021skeleton} & Three-stream & DGCNN & 88.0 & 94.9 & - & - \\
      Hi-TRS \cite{chen2022hierarchically} & Three-stream & Transformer & 90.0 & 95.7 & 85.3 & 87.4 \\
      \rowcolor{Gray!20} \multicolumn{7}{l}{\textit{Contrastive Learning:}} \\
      CPM \cite{zhang2022contrastive} & Single-stream & ST-GCN & 84.8 & 91.1 & 78.4 & 78.9 \\
      CrosSCLR \cite{li20213d} & Three-stream & ST-GCN & 86.2 & 92.5 & 80.5 & 80.4 \\
      AimCLR \cite{guo2022contrastive} & Three-stream & ST-GCN & 86.9 & 92.8 & 80.1 & 80.9 \\
      ActCLR \cite{lin2023actionlet} & Three-stream & ST-GCN & 88.2 & 93.9 & 82.1 & 84.6 \\
      HYSP \cite{franco2023hyperbolic} & Three-stream & ST-GCN & 89.1 & 95.2 & 84.5 & 86.3 \\
      \rowcolor{Gray!20} \multicolumn{7}{l}{\textit{Masked Prediction:}} \\
      SkeletonMAE \cite{wu2023skeletonmae} & Single-stream & STTFormer & 86.6 & 92.9 & 76.8 & 79.1 \\
      SkeletonMAE \cite{yan2023skeletonmae} & Single-stream & STRL & 92.8 & 96.5 & 84.8 & 85.7 \\
      MotionBERT \cite{zhu2023motionbert} & Single-stream & DSTformer & 93.0 & 97.2 & - & - \\
      MAMP \cite{mao2023masked} & Single-stream & Transformer & \textbf{93.1} & \underline{97.5} & \textbf{90.0} & \textbf{91.3} \\
      \midrule
      \textbf{Skeleton2vec(Ours)} & Single-stream & Transformer & \textbf{93.1} & \textbf{97.8} & \underline{89.5} & \underline{91.1} \\
      \bottomrule
    \end{tabular}
    \caption{
      Performance comparison in fine-tuning protocol on NTU 60 and NTU 120 datasets.
      The best results are shown in bold, and the second-best results
      are highlighted with an underline.
    }
    \label{tab:fine-tuning}
\end{table*}

\begin{table}[htbp]
    \centering
    \begin{tabular}{l c c c c}
      \toprule
      \multirow{3}{*}{Method} &
      \multicolumn{4}{c}{NTU 60} \\
      & \multicolumn{2}{c}{XSub(\%)} & \multicolumn{2}{c}{XView(\%)} \\
      & (1\%) & (10\%) & (1\%) & (10\%) \\
      \midrule
      LongTGAN \cite{zheng2018unsupervised} & 35.2 & 62.0 & - & - \\
      MS2L \cite{lin2020ms2l} & 33.1 & 65.1 & - & - \\
      ISC \cite{thoker2021skeleton} & 35.7 & 65.9 & 38.1 & 72.5 \\
      3s-CrosSCLR \cite{li20213d} & 51.1 & 74.4 & 50.0 & 77.8 \\
      3s-Colorization \cite{yang2021skeleton} & 48.3 & 71.7 & 52.5 & 78.9 \\
      3s-Hi-TRS \cite{chen2022hierarchically} & 49.3 & 77.7 & 51.5 & 81.1 \\
      3s-AimCLR \cite{guo2022contrastive} & 54.8 & 78.2 & 54.3 & 81.6 \\
      3s-CMD \cite{mao2022cmd} & 55.6 & 79.0 & 55.5 & 82.4 \\
      CPM \cite{zhang2022contrastive} & 56.7 & 73.0 & 57.5 & 77.1 \\
      SkeletonMAE \cite{wu2023skeletonmae} & 54.4 & 80.6 & 54.6 & 83.5 \\
      3s-HYSP \cite{franco2023hyperbolic} & - & 80.5 & - & 85.4 \\
      3s-SkeAttnCLR \cite{Hua2023SkeAttnCLR} & 59.6 & 81.5 & 59.2 & 83.8 \\
      MAMP \cite{mao2023masked} & 66.0 & 88.0 & 68.7 & 91.5 \\
      \midrule
      \textbf{Skeleton2vec(Ours)} & \textbf{75.7} & \textbf{89.2} & \textbf{76.2} & \textbf{92.9} \\
      \bottomrule
    \end{tabular}
    \caption{
      Performance comparison in the semi-supervised protocol on NTU 60 datasets.
      We averaged the results of five runs as the final performance.
    }
    \label{tab:semi-supervised}
\end{table}

\begin{table}[htbp]
    \centering
    \begin{tabular}{l c c}
      \toprule
      \multirow{2}{*}{Method} &
      \multicolumn{2}{c}{To PKU-II} \\
      & NTU 60 & NTU 120 \\
      \midrule
      LongTGAN \cite{zheng2018unsupervised} & 44.8 & - \\
      MS2L \cite{lin2020ms2l} & 45.8 & - \\
      ISC \cite{thoker2021skeleton} & 51.1 & 52.3 \\
      CMD \cite{mao2022cmd} & 56.0 & 57.0 \\
      HaLP+CMD \cite{shah2023halp} & 56.6 & 57.3 \\
      SkeletonMAE \cite{wu2023skeletonmae} & 58.4 & 61.0 \\
      MAMP \cite{mao2023masked} & 70.6 & 73.2 \\
      \midrule
      \textbf{Skeleton2vec(Ours)} & \textbf{73.0} & \textbf{75.1} \\
      \bottomrule
    \end{tabular}
    \caption{
      Performance comparison in the transfer learning protocol.
      The source datasets are NTU-60 and NTU-120, and the target dataset is PKU-II.
    }
    \label{tab:transer}
\end{table}

\noindent \textbf{Fine-tuning Evaluation:}
In the fine-tuning protocol, we add an MLP head to the pre-trained
encoder and then fine-tune the entire network. We use the AdamW optimizer with a
weight decay of 0.05. The learning rate starts at 0 and linearly increases to 3e-4
for the first 5 epochs, then decreases to 1e-5 according to a cosine decay schedule.
We train the network for a total of 100 epochs with a batch size of 48.
Evaluation of the fine-tuning results on the NTU-60 and NTU-120 datasets is
presented in \cref{tab:fine-tuning}. Our proposed Skeleton2vec consistently outperforms
previous methods based on the masked prediction task, including SkeletonMAE \cite{yan2023skeletonmae}
and MotionBERT \cite{zhu2023motionbert}, across all datasets. Moreover, our
approach demonstrates comparable results to the current state-of-the-art method,
MAMP \cite{mao2023masked}, and achieves further improvements on the NTU-60 XView dataset.

\noindent \textbf{Semi-supervised Evaluation:}
In the semi-supervised evaluation protocol, only 1\% and 10\% of the training
data are employed for fine-tuning, maintaining consistency with other training settings.
Evaluations on the NTU-60 dataset and comparisons with state-of-the-art approaches
such as HYSP \cite{franco2023hyperbolic}, SkeAttnCLR \cite{Hua2023SkeAttnCLR}, and MAMP
\cite{mao2023masked} are conducted.
As depicted in \cref{tab:semi-supervised},
Skeleton2vec demonstrates significant superiority over these methods, particularly
when utilizing only 1\% of the training data. Specifically, on the XSub and XView
settings, Skeleton2vec outperforms MAMP by 9.7\% and 7.5\%, respectively,
affirming the superiority of the proposed Skeleton2vec pretraining framework.

\noindent \textbf{Transfer Learning Evaluation:}
In the transfer learning evaluation protocol, pretraining is initially performed
on the source dataset and subsequently fine-tuned on the target dataset. The
source datasets used in our experiments are NTU-60 and NTU-120, with the target
dataset being PKU-MMD II. As illustrated in \cref{tab:transer}, our proposed
Skeleton2vec surpasses the state-of-the-art method MAMP by 2.4\% and 1.9\% when
using NTU-60 and NTU-120 as source datasets, respectively. This underscores the
robustness of features learned through the Skeleton2vec framework.

\subsection{Ablation Study}
We conducted an extensive ablation study on NTU-60 dataset to analyze the proposed
SKeleton2vec framework. Unless otherwise specified, we pre-train the model for 200
epochs and report the results under the linear evaluation protocol.

\begin{figure}
  \centering
  \includegraphics[width=0.95\linewidth]{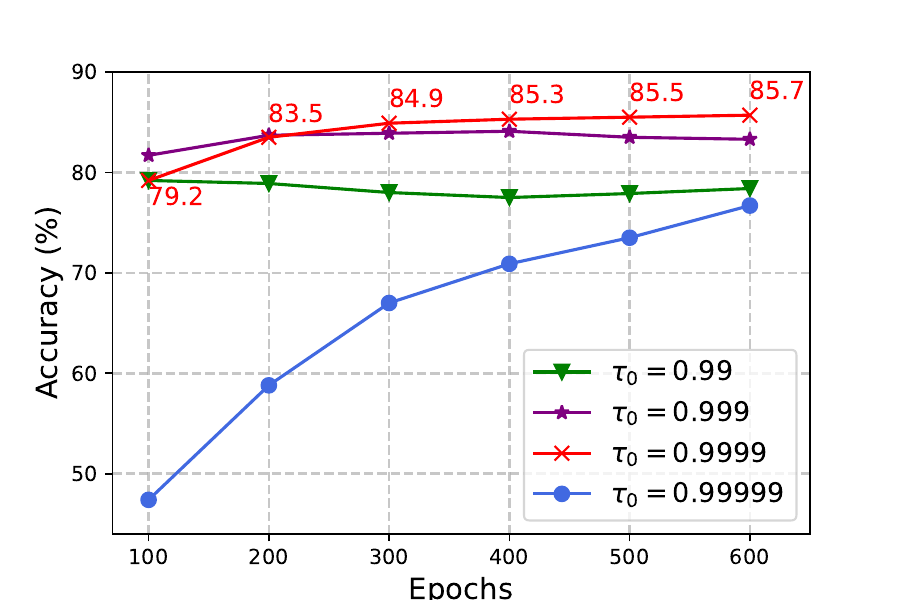}
  \caption{
    Ablation study on the EMA parameter $\tau_{0}$.
    The results are reported on the NTU-60 XSub dataset under the
    linear protocol.
  }
  \label{fig:ema_ablation}
\end{figure}

\noindent \textbf{Teacher Weight Update:}:
We regulate the update frequency of teacher's weights by adjusting the parameter
$\tau_{0}$ in the exponential moving average. In \cref{fig:ema_ablation}, we compared the
impact of four different values of $\tau_{0}$ on the pre-training performance of the model.
It is observed that employing smaller $\tau_{0}$ values (0.99, 0.999) leads to a rapid
performance improvement in the early stages of training (first 100 epochs). However,
as training progresses, the performance growth diminishes, and in some cases,
a decline is observed. Conversely, overly large values of $\tau_{0}$ (0.99999) significantly
slow down the convergence of training, incurring impractical time costs.
Through experimentation, we found that using an appropriate $\tau_{0}$ value (0.9999)
achieves a balanced convergence speed and growth potential,
resulting in optimal performance.

\begin{table}[htbp]
    \centering
    \begin{tabular}{c@{\hspace{6pt}}c c c c}
      \toprule
      \multirow{2}{*}{Strategy} &
      \multirow{2}{*}{$\alpha$} &
      \multirow{2}{*}{$\beta$} &
      \multicolumn{2}{c}{NTU 60} \\
      & & & XSub & XView \\
      \midrule
      Random masking & 1 & 0.0 & 79.4 & 85.1 \\
      Tube masking & 5 & 0.0 & 83.0 & 87.2 \\
      Motion-aware tube masking & 5 & 0.1 & \textbf{83.5} & \textbf{87.7} \\
      \bottomrule
    \end{tabular}
    \caption{
      Ablation study on the masking strategy.
      $\alpha$ represents the length of each tube, while $\beta$ denotes the
      parameter of motion-aware sampling.
    }
    \label{tab:masking_strategy}
\end{table}

\begin{table}[htbp]
    \centering
    \begin{subtable}[t]{0.495\linewidth}
      \begin{tabular}{c c c}
        \toprule
        \multirow{2}{*}{$\beta$} &
        \multicolumn{2}{c}{NTU 60} \\
        & XSub & XView \\
        \midrule
        0.0 & 83.0 & 87.2 \\
        0.1 & \textbf{83.5} & \textbf{87.7} \\
        0.2 & 82.1 & 87.0 \\
        0.3 & 79.5 & 86.3 \\
        \bottomrule
      \end{tabular}
      \caption{
        Motion-aware sampling
      }
      \label{tab:beta}
    \end{subtable}
    \begin{subtable}[t]{0.495\linewidth}
      \begin{tabular}{c c c}
        \toprule
        \multirow{2}{*}{$m$} &
        \multicolumn{2}{c}{NTU 60} \\
        & XSub & XView \\
        \midrule
        0.80 & 83.1 & 86.7 \\
        0.85 & 83.3 & 87.3 \\
        0.90 & \textbf{83.5} & \textbf{87.7} \\
        0.95 & 77.1 & 82.1 \\
        \bottomrule
      \end{tabular}
      \caption{
        Masking ratio
      }
      \label{tab:masking_ratio}
  \end{subtable}
  \caption{
    Ablation study on the masking ratio and motion-aware sampling.
  }
  \label{tab:masking_ratio_and_beta}
\end{table}

\begin{figure}[t]
  \centering
  \includegraphics[width=0.95\linewidth]{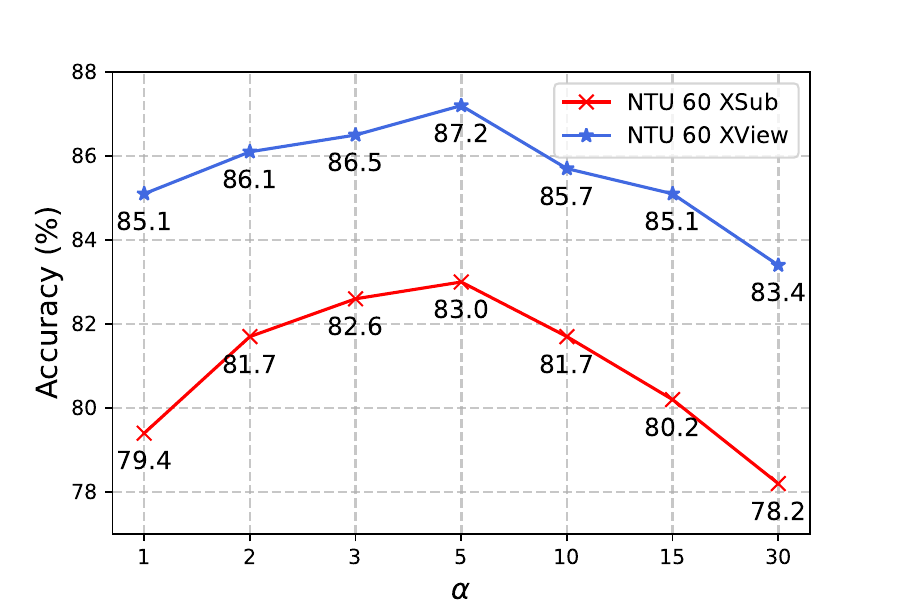}
  \caption{
    Ablation study on the tube length.
    $\alpha=0$ is equivalent to random masking, while $\alpha=30$, which is the length
    of the input sequence, is equivalent to single-tube masking.
  }
  \label{fig:alpha_ablation}
\end{figure}

\noindent \textbf{Masking Strategy:}
\cref{tab:masking_strategy} illustrates the effectiveness of our proposed
motion-aware tube masking strategy. We compared its performance with random masking
and tube masking (without motion-aware sampling). The results indicate a significant
performance boost with tube masking compared to random masking, showing improvements
of 3.6\% and 2.1\% under the XSub and XView testing protocols of the NTU-60 dataset,
respectively. This underscores the capability of tube segmentation to compel the
model into effective long-range motion modeling. Moreover, motion-aware tube masking
further improves performance, highlighting the value of guiding the model to focus
on semantically rich action regions. A detailed analysis of hyperparameters in
motion-aware tube masking will be presented in subsequent sections.

\noindent \textbf{Tube Length:}
We investigated the impact of the length $\alpha$ of each tube on pre-training performance.
As depicted in \cref{fig:alpha_ablation}, excessively short tube lengths result
in information leakage between adjacent frames, leading to a performance decline.
On the other hand, overly long tube lengths pose excessively challenging pre-training
tasks, impairing the model's learning capacity, as discussed in \cref{sec:motion-aware_tube_masking}.
Hence, selecting an appropriate tube length is crucial. Considering the results
from \cref{fig:alpha_ablation}, we identified a tube length of $\alpha=5$ as
optimal, achieving the best balance and performance.

\noindent \textbf{Motion-aware Sampling:}:
We compared the performance of learned representations under different motion-aware
sampling parameters $\beta$. As shown in \cref{tab:beta}, selecting an appropriate
sampling parameter enhances pre-training performance compared to not using
motion prior information ($\beta=0$). However, excessively large sampling parameters
can result in overly fixed sampling of joints, leading to a loss of diversity and
a subsequent performance decline. We empirically found that a sampling parameter of
$\beta=0.1$ yields the best results.

\noindent \textbf{Masking Ratio:}:
In \cref{tab:masking_ratio}, we compared the influence of different masking ratios
on the results. It is evident that excessively large or small masking ratios can
impair the final performance. We ultimately selected a masking ratio of 90\% to
achieve optimal results.
\subsection{Conclusion}
In this work, we propose Skeleton2vec, a novel self-supervised learning framework
for 3D skeleton-based action recognition. We demonstrated the superiority of utilizing
global contextualized representations built by a teacher model as the prediction target
for the masked prediction task, compared to isolated raw joints or temporal motion with
local context. Furthermore, considering the high spatiotemporal correlation in skeleton
sequences, we proposed the motion-aware tube masking strategy to compel the model into
effective long-range motion modeling. Extensive experiments conducted on three large-scale
prevalent benchmarks validated the effectiveness of our approach. The experimental results
showcased outstanding performance of our proposed Skeleton2vec,
achieving state-of-the-art results across multiple testing protocols.
{
    \small
    \bibliographystyle{ieeenat_fullname}
    \bibliography{main}

\begin{thebibliography}{49}
\providecommand{\natexlab}[1]{#1}
\providecommand{\url}[1]{\texttt{#1}}
\expandafter\ifx\csname urlstyle\endcsname\relax
  \providecommand{\doi}[1]{doi: #1}\else
  \providecommand{\doi}{doi: \begingroup \urlstyle{rm}\Url}\fi

\bibitem[Baevski et~al.(2022)Baevski, Hsu, Xu, Babu, Gu, and Auli]{baevski2022data2vec}
Alexei Baevski, Wei-Ning Hsu, Qiantong Xu, Arun Babu, Jiatao Gu, and Michael Auli.
\newblock Data2vec: A general framework for self-supervised learning in speech, vision and language.
\newblock In \emph{International Conference on Machine Learning}, pages 1298--1312. PMLR, 2022.

\bibitem[Baevski et~al.(2023)Baevski, Babu, Hsu, and Auli]{baevski2023efficient}
Alexei Baevski, Arun Babu, Wei-Ning Hsu, and Michael Auli.
\newblock Efficient self-supervised learning with contextualized target representations for vision, speech and language.
\newblock In \emph{International Conference on Machine Learning}, pages 1416--1429. PMLR, 2023.

\bibitem[Bao et~al.(2021)Bao, Dong, Piao, and Wei]{bao2021beit}
Hangbo Bao, Li Dong, Songhao Piao, and Furu Wei.
\newblock Beit: Bert pre-training of image transformers.
\newblock \emph{arXiv preprint arXiv:2106.08254}, 2021.

\bibitem[Cao et~al.(2018)Cao, Hidalgo, Simon, Wei, and Sheikh]{2018OpenPose}
Zhe Cao, Gines Hidalgo, Tomas Simon, Shih~En Wei, and Yaser Sheikh.
\newblock Openpose: Realtime multi-person 2d pose estimation using part affinity fields.
\newblock \emph{IEEE TPAMI}, 2018.

\bibitem[Chen et~al.(2021)Chen, Zhang, Yuan, Li, Deng, and Hu]{chen2021channel}
Yuxin Chen, Ziqi Zhang, Chunfeng Yuan, Bing Li, Ying Deng, and Weiming Hu.
\newblock Channel-wise topology refinement graph convolution for skeleton-based action recognition.
\newblock In \emph{ICCV}, pages 13359--13368, 2021.

\bibitem[Chen et~al.(2022)Chen, Zhao, Yuan, Tian, Xia, Geng, Han, and Metaxas]{chen2022hierarchically}
Yuxiao Chen, Long Zhao, Jianbo Yuan, Yu Tian, Zhaoyang Xia, Shijie Geng, Ligong Han, and Dimitris~N Metaxas.
\newblock Hierarchically self-supervised transformer for human skeleton representation learning.
\newblock In \emph{ECCV}, pages 185--202. Springer, 2022.

\bibitem[Dong et~al.(2023{\natexlab{a}})Dong, Sun, Liu, Chen, Liu, and Wang]{hico2023}
Jianfeng Dong, Shengkai Sun, Zhonglin Liu, Shujie Chen, Baolong Liu, and Xun Wang.
\newblock Hierarchical contrast for unsupervised skeleton-based action representation learning.
\newblock In \emph{AAAI}, 2023{\natexlab{a}}.

\bibitem[Dong et~al.(2023{\natexlab{b}})Dong, Bao, Zhang, Chen, Zhang, Yuan, Chen, Wen, Yu, and Guo]{dong2023peco}
Xiaoyi Dong, Jianmin Bao, Ting Zhang, Dongdong Chen, Weiming Zhang, Lu Yuan, Dong Chen, Fang Wen, Nenghai Yu, and Baining Guo.
\newblock Peco: Perceptual codebook for bert pre-training of vision transformers.
\newblock In \emph{AAAI}, pages 552--560, 2023{\natexlab{b}}.

\bibitem[Dosovitskiy et~al.(2020)Dosovitskiy, Beyer, Kolesnikov, Weissenborn, Zhai, Unterthiner, Dehghani, Minderer, Heigold, Gelly, et~al.]{dosovitskiy2020image}
Alexey Dosovitskiy, Lucas Beyer, Alexander Kolesnikov, Dirk Weissenborn, Xiaohua Zhai, Thomas Unterthiner, Mostafa Dehghani, Matthias Minderer, Georg Heigold, Sylvain Gelly, et~al.
\newblock An image is worth 16x16 words: Transformers for image recognition at scale.
\newblock \emph{arXiv preprint arXiv:2010.11929}, 2020.

\bibitem[Du et~al.(2015)Du, Fu, and Wang]{du2015skeleton}
Yong Du, Yun Fu, and Liang Wang.
\newblock Skeleton based action recognition with convolutional neural network.
\newblock In \emph{2015 3rd IAPR Asian conference on pattern recognition (ACPR)}, pages 579--583. IEEE, 2015.

\bibitem[Duan et~al.(2022)Duan, Wang, Chen, and Lin]{duan2022dg}
Haodong Duan, Jiaqi Wang, Kai Chen, and Dahua Lin.
\newblock Dg-stgcn: dynamic spatial-temporal modeling for skeleton-based action recognition.
\newblock \emph{arXiv preprint arXiv:2210.05895}, 2022.

\bibitem[Fang et~al.(2017)Fang, Xie, Tai, and Lu]{fang2017rmpe}
Hao-Shu Fang, Shuqin Xie, Yu-Wing Tai, and Cewu Lu.
\newblock Rmpe: Regional multi-person pose estimation.
\newblock In \emph{ICCV}, pages 2334--2343, 2017.

\bibitem[Franco et~al.(2023)Franco, Mandica, Munjal, and Galasso]{franco2023hyperbolic}
Luca Franco, Paolo Mandica, Bharti Munjal, and Fabio Galasso.
\newblock Hyperbolic self-paced learning for self-supervised skeleton-based action representations.
\newblock In \emph{Int. Conf. Learn. Represent.}, 2023.

\bibitem[Grill et~al.(2020)Grill, Strub, Altch{\'e}, Tallec, Richemond, Buchatskaya, Doersch, Avila~Pires, Guo, Gheshlaghi~Azar, et~al.]{grill2020bootstrap}
Jean-Bastien Grill, Florian Strub, Florent Altch{\'e}, Corentin Tallec, Pierre Richemond, Elena Buchatskaya, Carl Doersch, Bernardo Avila~Pires, Zhaohan Guo, Mohammad Gheshlaghi~Azar, et~al.
\newblock Bootstrap your own latent-a new approach to self-supervised learning.
\newblock \emph{NeurIPS}, 33:\penalty0 21271--21284, 2020.

\bibitem[Guo et~al.(2022)Guo, Liu, Chen, Liu, Wang, and Ding]{guo2022contrastive}
Tianyu Guo, Hong Liu, Zhan Chen, Mengyuan Liu, Tao Wang, and Runwei Ding.
\newblock Contrastive learning from extremely augmented skeleton sequences for self-supervised action recognition.
\newblock In \emph{AAAI}, pages 762--770, 2022.

\bibitem[He et~al.(2020)He, Fan, Wu, Xie, and Girshick]{he2020momentum}
Kaiming He, Haoqi Fan, Yuxin Wu, Saining Xie, and Ross Girshick.
\newblock Momentum contrast for unsupervised visual representation learning.
\newblock In \emph{CVPR}, pages 9729--9738, 2020.

\bibitem[He et~al.(2022)He, Chen, Xie, Li, Doll{\'a}r, and Girshick]{he2022masked}
Kaiming He, Xinlei Chen, Saining Xie, Yanghao Li, Piotr Doll{\'a}r, and Ross Girshick.
\newblock Masked autoencoders are scalable vision learners.
\newblock In \emph{CVPR}, pages 16000--16009, 2022.

\bibitem[Hua et~al.(2023)Hua, Wu, Zheng, Lu, Liu, Chen, and Wu]{Hua2023SkeAttnCLR}
Yilei Hua, Wenhan Wu, Ce Zheng, Aidong Lu, Mengyuan Liu, Chen Chen, and Shiqian Wu.
\newblock Part aware contrastive learning for self-supervised action recognition.
\newblock In \emph{Int. J. Comput. Vis.}, 2023.

\bibitem[Li et~al.(2017)Li, Zhong, Xie, and Pu]{li2017skeleton}
Chao Li, Qiaoyong Zhong, Di Xie, and Shiliang Pu.
\newblock Skeleton-based action recognition with convolutional neural networks.
\newblock In \emph{2017 IEEE international conference on multimedia \& expo workshops (ICMEW)}, pages 597--600. IEEE, 2017.

\bibitem[Li et~al.(2021)Li, Wang, Ni, Wang, Yang, and Zhang]{li20213d}
Linguo Li, Minsi Wang, Bingbing Ni, Hang Wang, Jiancheng Yang, and Wenjun Zhang.
\newblock 3d human action representation learning via cross-view consistency pursuit.
\newblock In \emph{CVPR}, pages 4741--4750, 2021.

\bibitem[Lin et~al.(2020)Lin, Song, Yang, and Liu]{lin2020ms2l}
Lilang Lin, Sijie Song, Wenhan Yang, and Jiaying Liu.
\newblock Ms2l: Multi-task self-supervised learning for skeleton based action recognition.
\newblock In \emph{ACM MM}, pages 2490--2498, 2020.

\bibitem[Lin et~al.(2023)Lin, Zhang, and Liu]{lin2023actionlet}
Lilang Lin, Jiahang Zhang, and Jiaying Liu.
\newblock Actionlet-dependent contrastive learning for unsupervised skeleton-based action recognition.
\newblock In \emph{CVPR}, pages 2363--2372, 2023.

\bibitem[Liu et~al.(2017)Liu, Hu, Li, Song, and Liu]{liu2017pku}
Chunhui Liu, Yueyu Hu, Yanghao Li, Sijie Song, and Jiaying Liu.
\newblock Pku-mmd: A large scale benchmark for continuous multi-modal human action understanding.
\newblock \emph{arXiv preprint arXiv:1703.07475}, 2017.

\bibitem[Liu et~al.(2016)Liu, Shahroudy, Xu, and Wang]{liu2016spatio}
Jun Liu, Amir Shahroudy, Dong Xu, and Gang Wang.
\newblock Spatio-temporal lstm with trust gates for 3d human action recognition.
\newblock In \emph{Computer Vision--ECCV 2016: 14th European Conference, Amsterdam, The Netherlands, October 11-14, 2016, Proceedings, Part III 14}, pages 816--833. Springer, 2016.

\bibitem[Liu et~al.(2019)Liu, Shahroudy, Perez, Wang, Duan, and Kot]{liu2019ntu}
Jun Liu, Amir Shahroudy, Mauricio Perez, Gang Wang, Ling-Yu Duan, and Alex~C Kot.
\newblock Ntu rgb+ d 120: A large-scale benchmark for 3d human activity understanding.
\newblock \emph{IEEE Trans. Pattern Anal. Mach. Intell.}, 42\penalty0 (10):\penalty0 2684--2701, 2019.

\bibitem[Mao et~al.(2022)Mao, Zhou, Lu, Deng, and Li]{mao2022cmd}
Yunyao Mao, Wengang Zhou, Zhenbo Lu, Jiajun Deng, and Houqiang Li.
\newblock Cmd: Self-supervised 3d action representation learning with cross-modal mutual distillation.
\newblock In \emph{ECCV}, pages 734--752. Springer, 2022.

\bibitem[Mao et~al.(2023)Mao, Deng, Zhou, Fang, Ouyang, and Li]{mao2023masked}
Yunyao Mao, Jiajun Deng, Wengang Zhou, Yao Fang, Wanli Ouyang, and Houqiang Li.
\newblock Masked motion predictors are strong 3d action representation learners.
\newblock In \emph{ICCV}, pages 10181--10191, 2023.

\bibitem[Moliner et~al.(2022)Moliner, Huang, and {\AA}str{\"o}m]{moliner2022bootstrapped}
Olivier Moliner, Sangxia Huang, and Kalle {\AA}str{\"o}m.
\newblock Bootstrapped representation learning for skeleton-based action recognition.
\newblock In \emph{CVPR}, pages 4154--4164, 2022.

\bibitem[Nie et~al.(2020)Nie, Liu, and Liu]{nie2020unsupervised}
Qiang Nie, Ziwei Liu, and Yunhui Liu.
\newblock Unsupervised 3d human pose representation with viewpoint and pose disentanglement.
\newblock In \emph{ECCV}, pages 102--118. Springer, 2020.

\bibitem[Rao et~al.(2021)Rao, Xu, Hu, Cheng, and Hu]{rao2021augmented}
Haocong Rao, Shihao Xu, Xiping Hu, Jun Cheng, and Bin Hu.
\newblock Augmented skeleton based contrastive action learning with momentum lstm for unsupervised action recognition.
\newblock \emph{Information Sciences}, 569:\penalty0 90--109, 2021.

\bibitem[Shah et~al.(2023)Shah, Roy, Shah, Mishra, Jacobs, Cherian, and Chellappa]{shah2023halp}
Anshul Shah, Aniket Roy, Ketul Shah, Shlok Mishra, David Jacobs, Anoop Cherian, and Rama Chellappa.
\newblock Halp: Hallucinating latent positives for skeleton-based self-supervised learning of actions.
\newblock In \emph{CVPR}, pages 18846--18856, 2023.

\bibitem[Shahroudy et~al.(2016)Shahroudy, Liu, Ng, and Wang]{shahroudy2016ntu}
Amir Shahroudy, Jun Liu, Tian-Tsong Ng, and Gang Wang.
\newblock Ntu rgb+ d: A large scale dataset for 3d human activity analysis.
\newblock In \emph{CVPR}, pages 1010--1019, 2016.

\bibitem[Su et~al.(2020)Su, Liu, and Shlizerman]{su2020predict}
Kun Su, Xiulong Liu, and Eli Shlizerman.
\newblock Predict \& cluster: Unsupervised skeleton based action recognition.
\newblock In \emph{CVPR}, pages 9631--9640, 2020.

\bibitem[Thoker et~al.(2021)Thoker, Doughty, and Snoek]{thoker2021skeleton}
Fida~Mohammad Thoker, Hazel Doughty, and Cees~GM Snoek.
\newblock Skeleton-contrastive 3d action representation learning.
\newblock In \emph{ACM MM}, pages 1655--1663, 2021.

\bibitem[Tong et~al.(2022)Tong, Song, Wang, and Wang]{tong2022videomae}
Zhan Tong, Yibing Song, Jue Wang, and Limin Wang.
\newblock Videomae: Masked autoencoders are data-efficient learners for self-supervised video pre-training.
\newblock \emph{NeurIPS}, 35:\penalty0 10078--10093, 2022.

\bibitem[Ulyanov et~al.(2016)Ulyanov, Vedaldi, and Lempitsky]{ulyanov2016instance}
Dmitry Ulyanov, Andrea Vedaldi, and Victor Lempitsky.
\newblock Instance normalization: The missing ingredient for fast stylization.
\newblock \emph{arXiv preprint arXiv:1607.08022}, 2016.

\bibitem[Vaswani et~al.(2017)Vaswani, Shazeer, Parmar, Uszkoreit, Jones, Gomez, Kaiser, and Polosukhin]{vaswani2017attention}
Ashish Vaswani, Noam Shazeer, Niki Parmar, Jakob Uszkoreit, Llion Jones, Aidan~N Gomez, {\L}ukasz Kaiser, and Illia Polosukhin.
\newblock Attention is all you need.
\newblock \emph{NeurIPS}, 30, 2017.

\bibitem[Wei et~al.(2022)Wei, Fan, Xie, Wu, Yuille, and Feichtenhofer]{wei2022masked}
Chen Wei, Haoqi Fan, Saining Xie, Chao-Yuan Wu, Alan Yuille, and Christoph Feichtenhofer.
\newblock Masked feature prediction for self-supervised visual pre-training.
\newblock In \emph{CVPR}, pages 14668--14678, 2022.

\bibitem[Wu et~al.(2023)Wu, Hua, Zheng, Wu, Chen, and Lu]{wu2023skeletonmae}
Wenhan Wu, Yilei Hua, Ce Zheng, Shiqian Wu, Chen Chen, and Aidong Lu.
\newblock Skeletonmae: Spatial-temporal masked autoencoders for self-supervised skeleton action recognition.
\newblock In \emph{2023 IEEE International Conference on Multimedia and Expo Workshops (ICMEW)}, pages 224--229. IEEE, 2023.

\bibitem[Xie et~al.(2022)Xie, Zhang, Cao, Lin, Bao, Yao, Dai, and Hu]{xie2022simmim}
Zhenda Xie, Zheng Zhang, Yue Cao, Yutong Lin, Jianmin Bao, Zhuliang Yao, Qi Dai, and Han Hu.
\newblock Simmim: A simple framework for masked image modeling.
\newblock In \emph{CVPR}, pages 9653--9663, 2022.

\bibitem[Xu et~al.(2020)Xu, Yu, Ni, Yang, Yang, and Zhang]{xu2020deep}
Jingwei Xu, Zhenbo Yu, Bingbing Ni, Jiancheng Yang, Xiaokang Yang, and Wenjun Zhang.
\newblock Deep kinematics analysis for monocular 3d human pose estimation.
\newblock In \emph{CVPR}, pages 899--908, 2020.

\bibitem[Yan et~al.(2023)Yan, Liu, Wei, Li, Li, and Lin]{yan2023skeletonmae}
Hong Yan, Yang Liu, Yushen Wei, Zhen Li, Guanbin Li, and Liang Lin.
\newblock Skeletonmae: graph-based masked autoencoder for skeleton sequence pre-training.
\newblock In \emph{ICCV}, pages 5606--5618, 2023.

\bibitem[Yan et~al.(2018)Yan, Xiong, and Lin]{yan2018spatial}
Sijie Yan, Yuanjun Xiong, and Dahua Lin.
\newblock Spatial temporal graph convolutional networks for skeleton-based action recognition.
\newblock In \emph{AAAI}, 2018.

\bibitem[Yang et~al.(2021)Yang, Liu, Lu, Er, and Kot]{yang2021skeleton}
Siyuan Yang, Jun Liu, Shijian Lu, Meng~Hwa Er, and Alex~C Kot.
\newblock Skeleton cloud colorization for unsupervised 3d action representation learning.
\newblock In \emph{ICCV}, pages 13423--13433, 2021.

\bibitem[Zhang et~al.(2022)Zhang, Hou, Zhang, and Li]{zhang2022contrastive}
Haoyuan Zhang, Yonghong Hou, Wenjing Zhang, and Wanqing Li.
\newblock Contrastive positive mining for unsupervised 3d action representation learning.
\newblock In \emph{ECCV}, pages 36--51. Springer, 2022.

\bibitem[Zhang et~al.(2017)Zhang, Lan, Xing, Zeng, Xue, and Zheng]{zhang2017view}
Pengfei Zhang, Cuiling Lan, Junliang Xing, Wenjun Zeng, Jianru Xue, and Nanning Zheng.
\newblock View adaptive recurrent neural networks for high performance human action recognition from skeleton data.
\newblock In \emph{Proceedings of the IEEE international conference on computer vision}, pages 2117--2126, 2017.

\bibitem[Zheng et~al.(2018)Zheng, Wen, Liu, Long, Dai, and Gong]{zheng2018unsupervised}
Nenggan Zheng, Jun Wen, Risheng Liu, Liangqu Long, Jianhua Dai, and Zhefeng Gong.
\newblock Unsupervised representation learning with long-term dynamics for skeleton based action recognition.
\newblock In \emph{AAAI}, 2018.

\bibitem[Zhou et~al.(2023)Zhou, Duan, Rao, Su, and Wang]{Zhou2023SelfsupervisedAR}
Yujie Zhou, Haodong Duan, Anyi Rao, Bing Su, and Jiaqi Wang.
\newblock Self-supervised action representation learning from partial spatio-temporal skeleton sequences.
\newblock In \emph{AAAI}, 2023.

\bibitem[Zhu et~al.(2023)Zhu, Ma, Liu, Liu, Wu, and Wang]{zhu2023motionbert}
Wentao Zhu, Xiaoxuan Ma, Zhaoyang Liu, Libin Liu, Wayne Wu, and Yizhou Wang.
\newblock Motionbert: A unified perspective on learning human motion representations.
\newblock In \emph{ICCV}, pages 15085--15099, 2023.

\end{thebibliography}
}


\end{document}